%% file: root.tex
\documentclass[letterpaper, 10 pt, conference]{ieeeconf}  % Comment this line out if you need a4paper

\IEEEoverridecommandlockouts                              % This command is only needed if 
\usepackage{times}

\usepackage{multicol}
\usepackage[bookmarks=true,hidelinks,colorlinks=true,urlcolor=blue,citecolor=blue,allcolors=blue]{hyperref}
\usepackage{caption}
\input{preamble}

\author{Yi Du$^{1*}$, Taimeng Fu$^{1*}$, Zhipeng Zhao$^{1}$, Shaoshu Su$^{1}$, Zitong Zhan$^{1}$, Qiwei Du$^{1}$,\\
Zhuoqun Chen$^{1}$, Bowen Li$^{2}$, Chen Wang$^{1}$\\
\texttt{\url{https://sairlab.org/vlnav/}}
\vspace{-10pt}
\thanks{$^{*}$Equal contribution.}%
\thanks{$^{1}$Spatial AI \& Robotics Lab, University at Buffalo, USA.}%
\thanks{$^{2}$Robotics Institute, Carnegie Mellon University, USA.}%
}

\DeclareCaptionFont{Smallfont}{\small}
\begin{document}
\title{\LARGE \bf
VL-Nav: Neuro-Symbolic Reasoning-based Vision-Language Navigation
\vspace{-10pt}
}

\maketitle
\thispagestyle{empty}
\pagestyle{empty}

%%%%%%%%%%%%%%%%%%%%%%%%%%%%%%%%%%%%%%%%%%%%%%%%%%%%%%%%%%%%%%%%%%%%%%%%%%%%%%%%

\input{sec/0_abstract}
\input{sec/1_intro}
\input{sec/2_related}
\input{sec/3_method}

\input{sec/4_evaluation}
% \input{sec/5_discussion}
\input{sec/6_conclusion}

\section*{Acknowledgments}
This work was partially supported by the Sony Faculty Innovation Award and the DARPA award HR00112490426. Any opinions, findings, conclusions, or recommendations expressed in this paper are those of the authors and do not necessarily reflect the views of DARPA or Sony.
% \vspace{-2pt}

\bibliographystyle{IEEEtran}
\bibliography{references}

\end{document}

%% file: preamble.tex
\usepackage[export]{adjustbox}
\usepackage{graphicx}   % for \includegraphics
\usepackage{subcaption}

\usepackage{float} 
\usepackage{placeins}

\usepackage{algorithm}
\usepackage{algorithmic}

\usepackage{booktabs}
\usepackage{utfsym}

\usepackage{wrapfig}
\usepackage{multirow}
\usepackage{multicol}
\usepackage{amsmath}
\usepackage{nicefrac}
\usepackage{amssymb}
\usepackage{bm}
\renewcommand{\paragraph}[1]{\noindent\textbf{#1}}
\newcommand{\fref}[1]{Fig.~\ref{#1}}

\newcommand{\eref}[1]{Equation~(\ref{#1})}

%% file: sec/0_abstract.tex
\begin{abstract}
% \label{sec:abstract}
Navigating unseen, large-scale environments based on complex and abstract human instructions remains a formidable challenge for autonomous mobile robots. 
Addressing this requires robots to infer implicit semantics and efficiently explore large-scale task spaces.
However, existing methods, ranging from end-to-end learning to foundation model-based modular architectures, often lack the capability to decompose complex tasks or employ efficient exploration strategies, leading to robot aimless wandering or target recognition failures. 
To address these limitations, we propose \textbf{VL-Nav}, a neuro-symbolic (NeSy) vision-language navigation system. The proposed system intertwines neural reasoning with symbolic guidance through two core components: (1) a \textbf{NeSy task planner} that leverages a symbolic 3D scene graph and image memory system to enhance the vision-language models' (VLMs) neural reasoning capabilities for task decomposition and replanning; 
and (2) a \textbf{NeSy exploration system} that couples neural semantic cues with the symbolic heuristic function to efficiently gather the task-related information while minimizing unnecessary repeat travel during exploration. 
Validated on the DARPA TIAMAT Challenge navigation tasks, our system achieved an 83.4\% success rate (SR) in indoor environments and 75\% in outdoor scenarios. 
VL-Nav achieved an 86.3\% SR in real-world experiments, including a challenging 483-meter run. 
Finally, we validate the system with complex instructions in a 3D multi-floor scenario. 
% Project Website: \url{https://sairlab.org/vlnav/}
\end{abstract}

\IEEEpeerreviewmaketitle

%% file: sec/1_intro.tex
% \vspace{-5pt}
\section{Introduction}
\label{sec:intro}

% \vspace{-5pt}
Efficient navigation following complex human instructions in completely unseen environments remains a formidable challenge for autonomous mobile robots. For example, consider a navigation task in the DARPA TIAMAT challenge \cite{noorani2025abstraction}: ``Today's weather report indicates rain. Help Rob find an umbrella, an appropriate jacket, and shoes.'' This marks a critical shift from naive command following vision-and-language navigaion \cite{gu2022vision} to reasoning-based vision-language navigation (VLN), where the robot must bridge massive logical gaps. The robot must first accurately understand the implicit semantics of the \textbf{abstract} instruction, reasoning out semantic inferences (e.g., ``rain'' implies waterproof gear) and performing object disambiguation to identify that the required targets are an umbrella, a rain jacket, and rain boots. This is essential to prevent the robot from looking for a normal jacket or sneakers as the target objects. Subsequently, it must efficiently explore the unknown environment to find all these required items within a limited task time. 

\begin{figure}[t]
    \centering
    \includegraphics[width=1\linewidth]{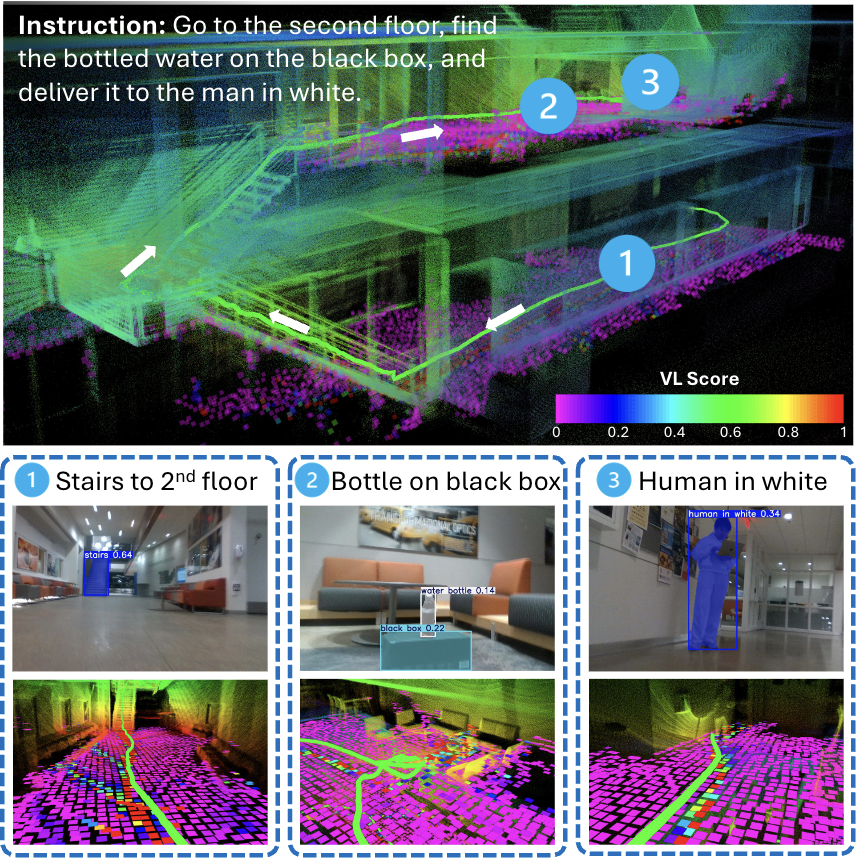} 
    % \vspace{-20pt}
    \caption{Given the complex instruction, VL-Nav autonomously navigates a university building. Top: The generated 3D global map and robot trajectory (green line). The projected value map (rainbow squares on the floor) visualizes the NeSy scoring policy, which prioritizes task-relevant areas to guide efficient exploration. Bottom: Key reasoning milestones where the system (1) identifies stairs for floor transition, (2) grounds the target object using spatial relation (on black box), and (3) verifies the recipient based on visual attributes.}
    \label{fig:head}
    \vspace{-15pt}
\end{figure}

However, a significant gap remains in addressing such complex VLN. Classical semantic-free approaches \cite{elfes1989using,fennema1990experiments,yamauchi1997frontier} lack the linguistic reasoning necessary to comprehend abstract human instructions.
Conversely, end-to-end learning methods, such as RL-based \cite{mirowski2016learning,savva2017minos,codevilla2018end} and VLA-based approaches \cite{zhang2024uni,cheng2024navila,zhang2025embodied}, are notoriously data-hungry and struggle with sim-to-real transfer. While recent foundation model-based modular architectures \cite{yokoyama2024vlfm, gadre2023cows, yin2025sg, zhang2025apexnav} integrate advanced VLMs for zero-shot single-object navigation, their tight coupling of target verification with exploration constrains inferential capabilities, frequently resulting in the identification of incorrect items (e.g., random shoes instead of rain boots). Furthermore, most of these methods over-rely on neural semantic cues while neglecting the potential of geometric frontiers, leading to aimless wandering within explored regions.
Consequently, existing approaches fundamentally lack two critical capabilities: (1) \textit{Accurate Task Reasoning} to parse and decompose abstract multi-target instructions; and (2) \textit{Efficient Exploration Strategies} to rapidly locate multiple targets in large-scale environments. 

To seamlessly bridge this gap, we propose a NeSy vision-language navigation system that fundamentally intertwines neural reasoning with symbolic guidance. Specifically, our VL-Nav architecture consists of two main modules:
(1) \textbf{NeSy Task Planner:} This planner logically decomposes abstract, multi-target instructions into atomic subtasks, dynamically issuing either \textit{exploration} subtasks to gather information or \textit{go to} subtasks to navigate to and report target objects. Crucially, the planner grounds VLM's neural reasoning in a unified symbolic memory comprising a 3D scene graph and object-centric image memory, anchoring VLM decisions in precise spatial awareness and historical context.
(2) \textbf{NeSy Exploration System:} Serving as the exploration backbone, this module couples neural semantic cues from lightweight VLMs with a symbolic heuristic function. This integration rapidly guides the robot through unknown environments to discover targets while minimizing unnecessary travel.

We comprehensively validate our proposed VL-Nav system in both simulations and extensive real-world robot environments. When evaluated in the navigation tasks in the DARPA TIAMAT challenge, our system demonstrated remarkable performance, achieving an 83.4\% success rate (SR) in indoor settings and 75\% in outdoor environments. 
Furthermore, real-world deployment across four diverse environments confirms the capability of generalizing from highly occluded, indoor domains to unstructured outdoor environments. Our real-world experiments achieved an 86.3\% success rate, covering challenging scenarios such as a 483-meter long-range trajectory. Additionally, we showcased a complex multi-floor demonstration as shown in \fref{fig:head}.

Our main contributions are as follows:
\begin{itemize}
\item We introduce VL-Nav, a NeSy system designed to resolve reasoning-based VLN tasks by intertwining neural semantic understanding with symbolic guidance.
\item We introduce a robust NeSy Task Planner to guide VLM reasoning and replanning within a unified symbolic memory. When coupled with our NeSy Exploration System, this approach facilitates rapid multi-target discovery and minimizes unnecessary travel.
\item VL-Nav achieves \textbf{83.4\%} and \textbf{75\%} SR in the indoor and outdoor settings on the navigation tasks of the DARPA TIAMAT Challenge, respectively. In real-world experiments, it attains an \textbf{86.3\%} SR, including long-range runs of up to \textbf{483 meters} and demonstrations of complex tasks in \textbf{multi-floor} environments.
\end{itemize}

%% file: sec/2_related.tex
\section{Related Works}
\label{sec:related}

Robot navigation methods can be broadly categorized into classical, end-to-end learning, and modular learning methods.

\paragraph{Classical Approaches:}
Classical methods for navigation, including Simultaneous Localization and Mapping (SLAM)-based techniques, have been extensively studied over the last three decades \cite{elfes1989using,fennema1990experiments,thrun1999minerva,yamauchi1997frontier}. 
These methods typically build geometric maps of an environment using depth sensors or monocular RGB cameras \cite{thrun2001robust,newcombe2011kinectfusion,davison2007monoslam,sattler2018benchmarking}, while simultaneously localizing the robot relative to the growing map. 
Exploration is often guided by heuristic strategies like frontier-based methods \cite{yamauchi1997frontier,freda2005frontier}, while analytical planners are employed for obstacle avoidance and path planning. 
Although effective for traditional navigation tasks, these classical methods lack the ability to integrate vision-language features, making them insufficient for human-collaboration tasks like vision-language navigation.

\paragraph{End-to-End Learning:}
End-to-end learning methods seek to directly map sensory inputs and instructions to navigation actions. Traditionally, these approaches relied on deep neural networks trained via imitation learning (IL) or reinforcement learning (RL) \cite{mirowski2016learning,savva2017minos,codevilla2018end,singh2023scene,ramrakhya2022habitat}. More recently, Vision-Language-Action (VLA) models \cite{zhang2024uni,cheng2024navila,zhang2025embodied} have emerged, attempting to unify vision, language, and continuous control into single massive transformer architectures. 
Despite their ability to inherently learn semantic priors for goal-directed exploration, these end-to-end approaches suffer from significant limitations when applied to real-world, multi-target VLN. 
They are notoriously data-hungry, computationally expensive to run in real-time, and lack interpretability. Most critically, they frequently fail to generalize beyond their training distributions, leading to severe performance degradation during sim2real transfer \cite{gervet2023navigating}.

\paragraph{Modular Learning:}
Modular learning approaches seek to combine the advantages of classical and end-to-end methods by replacing specific components of the classical pipeline with learned modules. This modularity allows for the integration of semantic priors while maintaining system interpretability and efficiency \cite{mcallister2017concrete,muller2018driving,chaplot2020learning,chaplot2007object}. For example, semantic maps generated from RGB-D data have been used to guide navigation toward specific objects \cite{chaplot2020object,ramakrishnan2020occupancy,ramakrishnan2022poni,hahn2021no}. Modular methods have shown promise in addressing Sim-to-Real transfer challenges by abstracting raw sensor data into higher-level representations, mitigating the impact of domain gaps \cite{muller2018driving, chaplot2007object, chaplot2020learning}. However, unlike neuro-symbolic methods like imperative learning \cite{wang2025imperative} existing modular approaches often rely heavily on real-world training data and struggle to utilize vision-language features.

\paragraph{Foundation Model-Based Modular Learning Methods:}
Recent advances in foundation models like large Vision-Language Models (VLMs) and Large Language Models (LLMs) have further enhanced semantic navigation by incorporating natural language understanding into modular learning approaches \cite{gadre2023cows, yin2025sg, yokoyama2024vlfm,zhang2025apexnav}. 
For instance, the Vision-Language Frontier Maps (VLFM) method \cite{yokoyama2024vlfm} leverages a pre-trained VLM to extract semantic values directly from RGB images, enabling zero-shot semantic prior understanding for frontier-based exploration. 
Other methods like ApexNav \cite{zhang2025apexnav} and SG-Nav \cite{yin2025sg} build open-vocabulary representations to find novel categories. However, these systems inherently operate under the premise of naive instruction following---where the ``target'' is explicitly named (e.g., ``go to the L-shape sofa''). They lack the complex logic required for Reasoning-based VLN needed in abstract and multi-target tasks where explicit targets are not provided (e.g., inferring that ``rain'' requires an ``rain jacket'' instead of a random coat). Furthermore, executing heavy foundation models all the time during the navigation process bottlenecks real-time operation on resource-constrained platforms.

%% file: sec/3_method.tex
\section{Methodology}
\label{sec:method}

\begin{figure}[h]
    \centering
    \includegraphics[width=1\linewidth]{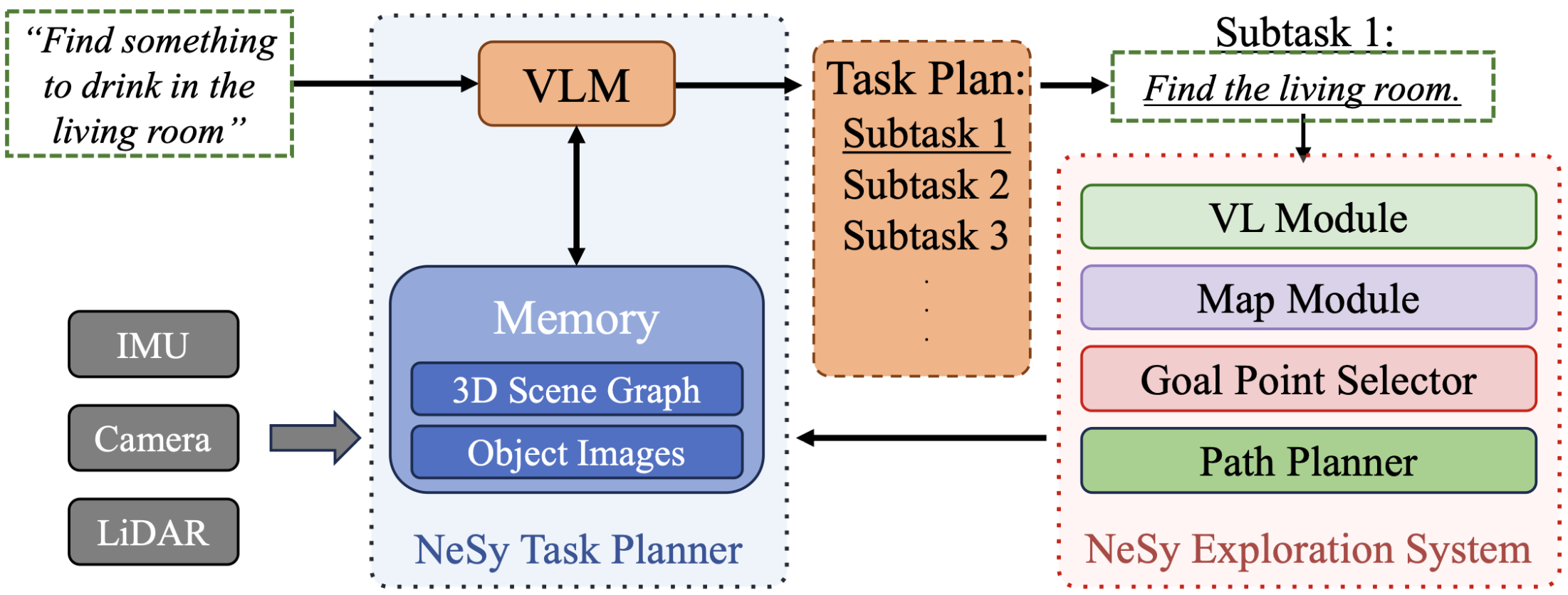} 
    % \vspace{-20pt}
    \caption{System pipeline overview.Complex tasks are decomposed into subtasks via VLM reasoning, guided by the unified memory system. The first subtask always serves as input to the NeSy exploration system, which either guides exploration or sets a direct goal point for robot navigation.}
    \label{fig:system_pipeline}
    \vspace{-10pt}
\end{figure}
To address the limitations of existing methods in resolving abstract and multi-target tasks, we introduce \textbf{VL-Nav}, a neuro-symbolic vision-language navigation system that intertwines neural semantic understanding with symbolic precision. As shown in \fref{fig:system_pipeline}, the architecture consists:
\begin{enumerate}
\item \textbf{NeSy Task Planner} logically decomposes complex, abstract instructions into atomic subtasks, \textit{exploration}  and \textit{go to}, by guiding a VLM's neural reasoning with a unified symbolic memory comprising a 3D scene graph and an object-centric image memory for task decomposition, replanning, and completion determination.
\item \textbf{NeSy Exploration System} couples neural semantic cues from lightweight open-vocabulary detection models (OVDMs) with a symbolic heuristic function, enabling efficient task-guided exploration of unknown environments without the computational bottlenecks of pure VLM-in-the-loop approaches.
\end{enumerate}

\subsection{NeSy Task Planner}

\subsubsection{Unified Memory System}
The Unified Memory System consists of a 3D scene graph and an object-centric image memory. First, we construct a 3D scene graph comprising two types of nodes: object nodes and room nodes, inspired by Hydra \cite{hughes2024foundations}. 
Room node masks are generated using our room segmentation algorithm, which relies on morphological operations. To label these room nodes, we utilize a Large Language Model (LLM) that infers labels based on the object nodes connected to each room node. The edges between object and room nodes are established through spatial inclusion; specifically, an edge is created if an object's centroid falls within a room's segmentation mask. 

Object nodes are generated by an open-vocabulary detector. Each object node stores four key attributes: (1) the object centroid; (2) the highest-confidence detection score; (3) the robot's pose at the viewpoint of this detection; and (4) the corresponding best-viewpoint RGB image. This integration of the 3D scene graph with object-centric best-viewpoint image memory forms a \textit{unified symbolic memory}, providing the spatial and relational grounding necessary for subsequent task decomposition and replanning.

\subsubsection{Task Decomposition and Replanning}
Leveraging Qwen3-VL \cite{bai2025qwen3} as the VLM backbone, the planner decomposes complex task descriptions into atomic ``exploration'' and ``go to'' subtasks. The VLM maintains a dynamic target object list, querying the memory module to track and register potential candidates. Upon the completion of each subtask, a replanning phase will be triggered, generating updated subtasks based on the updated symbolic memory.

For target acquisition, the system employs a coarse-to-fine verification strategy. For instance, when locating a specific item like a ``rain jacket,'' the ``go to'' subtask is determined via a two-step process: \textbf{(1) Symbolic Filtering:} The 3D scene graph filters the available object nodes to propose the top-$k$ (e.g., top-3) candidates based on detection confidence. \textbf{(2) Neural Verification:} The VLM performs fine-grained reasoning on the saved best-view images and the neighbor scene graph nodes of these top candidates to determine which object best semantically matches the abstract instruction.
Once the target is chosen, the planner publishes the saved best-view pose as the navigation goal, directing the exploration system to navigate to and report the identified target.

\begin{figure*}[t]
    \centering
    \includegraphics[width=0.86\linewidth]{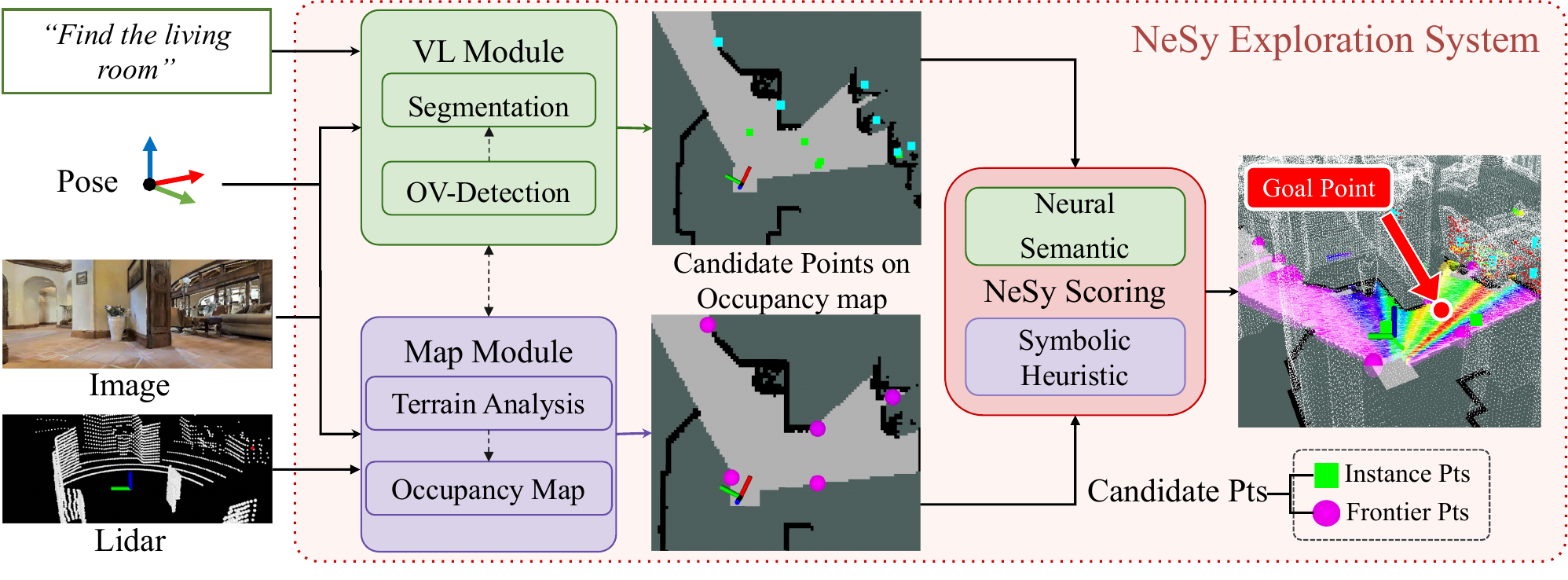} 
    % \vspace{-20pt}
    % \captionsetup{
    % width=\textwidth,
    % font=Smallfont,
    % labelfont=Smallfont,
    % textfont=Smallfont
    % }
    \caption{An overview of the NeSy exploration system. It processes inputs including prompts, RGB images, poses, and LiDAR scans. The Vision-Language (VL) module conducts open-vocabulary detection to identify objects related to the prompt, generating instance-based target points. Concurrently, the map module performs terrain analysis and manages a dynamic occupancy map. Frontier-based target points are then identified based on this occupancy map, along with the instance points, forming a candidate points pool. Finally, this system employs the NeSy scoring to select the most effective goal point.}
    \label{fig:pipeline}
    \vspace{-6pt}
\end{figure*}

\begin{figure}[t]
    \centering
    \includegraphics[width=1\linewidth]{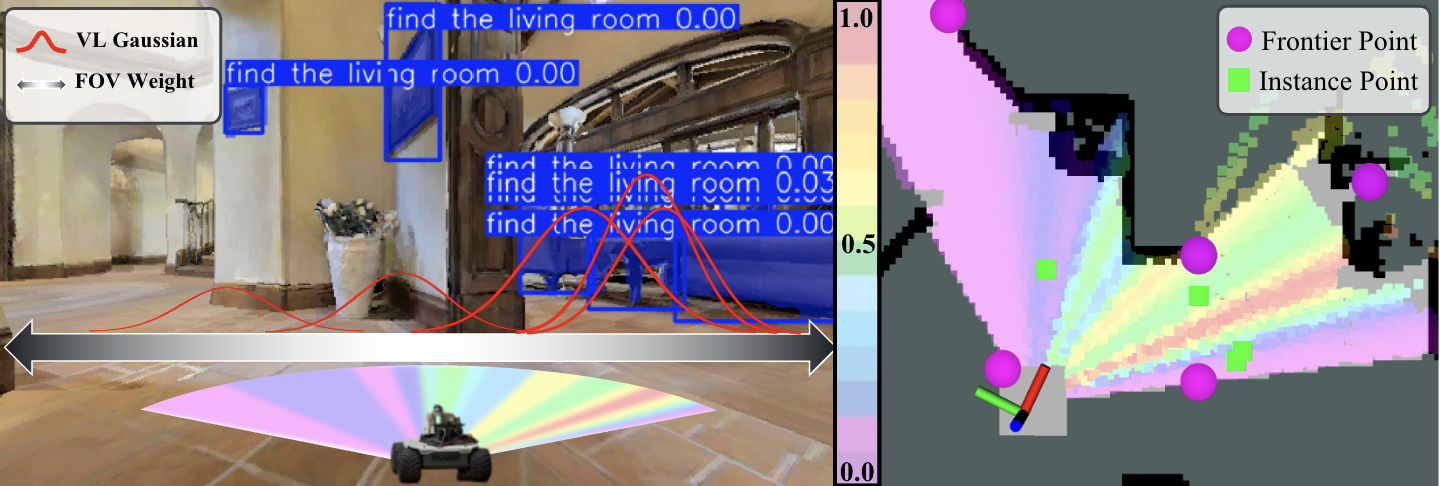} 
    % \vspace{-20pt}
    \caption{A brief illustration of VL Scoring. The pixel-wise open-vocabulary detection results are transferred into the spatial distibution via the Gaussian mixture model regularized by the FoV weighting (the gray arrows). Then the frontier-based and the instance-based target points will be assigned with VL score based on the distribution \eref{eq:VL}.}
    \label{fig:NeSy}
    \vspace{-13pt}
\end{figure}

\subsection{NeSy Exploration System}
Taking the subtask from the task planner, the \textbf{NeSy Exploration System} operationalizes high-level directives by fusing neural semantic cues from open-vocabulary detectors with symbolic geometric heuristics. The pipeline as shown in \fref{fig:pipeline} proceeds from the generation of candidate goal points to the hybrid NeSy scoring policy, facilitating rapid, task-guided exploration and navigation in real-time.
\subsubsection{Frontier-based Target Points}

A cell $(m_x,m_y)$ is a \emph{frontier} if it is free, $\mathcal{G}(m_x,m_y)=0$, and at least one neighbor is unknown, $\mathcal{G}(m_x+\Delta x,m_y+\Delta y)=-1$ for some $(\Delta x,\Delta y)$. After every grid update, only cells in the forward FoV wedge are tested:
\begin{equation}\label{eq:frontier}
\resizebox{0.66\linewidth}{!}{$
\begin{split}
\Bigl|\mathrm{Angle}\bigl((m_x,m_y),(x_r,y_r)\bigr) - \theta_r\Bigr|
\;&\le\; \frac{\mathrm{hfov}}{2}, \\
\text{and} \quad \|(m_x,m_y) - (x_r,y_r)\|
\;&\le\; R,
\end{split}
$}
\end{equation}
where $(x_r,y_r)$ and $\theta_r$ denote the robot's current position and heading, $\mathrm{hfov}$ represents the horizontal field of view, and $R$ is the maximum effective sensor range.

A breadth-first search (BFS) clusters these frontier cells. Each cluster is represented by either a single centroid (for small clusters) or multiple sampled points (for large clusters) as the frontier-based target points. 

\subsubsection{Instance-Based Target Points (IBTP)}

The vision-language detector shown in \fref{fig:pipeline} VL Module periodically reports \emph{candidate instance centers} in the form $(q_x,\;q_y,\;\text{confidence})$,
where \(q_x, q_y\) denote the estimated global coordinates for a potential target instance, and \(\text{confidence}\) quantifies how likely this detection is to match the desired instance. If \(\text{confidence is larger than the detection threshold } \tau_{\mathrm{det}}\), the candidate is retained; otherwise, it is discarded as too uncertain. Retained points will be downsampled via a voxel-grid filter if many candidates lie in close proximity.

This \emph{instance-based} approach mimics human behavior when searching for an object: upon glimpsing something that \emph{might} match the target, one naturally moves closer to confirm. VL-Nav likewise does not ignore intermediate detections. Instead, any candidate above a confidence threshold is treated as a valid \emph{goal candidate}, allowing the robot to approach and verify whether it truly is an instance of interest. If the detection turns out to be incorrect, the robot continues exploring via frontiers or other instance cues.

\subsubsection{NeSy Scoring Policy}

Once frontier-based and instance-based target points are gathered as candidate goals, the system computes the NeSy score for each candidate goal \(\mathbf{g}\). Let \((x_r,y_r)\) be the robot position, \(\theta_r\) the heading, and \(d(\mathbf{x}_r,\mathbf{g})\) the Euclidean distance. The angular offset is:
\begin{equation}\label{eq:offset}
\resizebox{0.58\linewidth}{!}{$
\Delta\theta
\;=\;
\mathrm{Angle}\bigl(\mathbf{g},(x_r,y_r)\bigr)
\;-\;\theta_r.
$}
\end{equation}

\paragraph{VL Score:}
As shown in \fref{fig:NeSy}, a vision-language (VL) score \(S_{\mathrm{VL}}(\mathbf{g})\) translates \emph{fine-grained vision-language features} from the YOLO-World \cite{cheng2024yolo} and FastSAM \cite{zhao2023fast} models into a Gaussian-mixture distribution over the robot’s horizontal field of view (FoV). Suppose the open-vocabulary detection model identifies \(K\) \emph{likely directions} or \emph{detection regions}, each parameterized by \(\bigl(\mu_k,\sigma_k,\alpha_k\bigr)\). Here, \(\mu_k\) indicates the mean offset angle within the FoV, \(\sigma_k\) encodes the detection’s angular uncertainty (in our implementation, a fixed constant of \(0.1\) for all detections), and \(\alpha_k\) is a confidence-based weight that captures how important that region is. By accumulating each region’s contribution in a mixture model, \(S_{\mathrm{VL}}(\mathbf{g})\) naturally biases the robot toward goals in the direction of high-confidence detections that match the subtask prompt.

From a human perspective, this process resembles \emph{observing} different areas that related to the target instance with varying levels of certainty. For instance, if someone is looking for a ``red chair'' in a large room, they might spot multiple \emph{red silhouettes} at the right side of their vision, plus a clearer view of something red near the center. Although peripheral glimpses (``out of the corner of the eye'') often carry lower confidence, they are not ignored. Instead, we naturally merge these partial observations into a mental sense of ``the chair is probably in one of these directions.'' The system follows a similar approach, assigning each detection region a Gaussian shape around \(\mu_k\), scaled by \(\sigma_k\) and weighted by \(\alpha_k\). Inspired by VLFM \cite{yokoyama2024vlfm}, we then multiply by \(C(\Delta\theta)\), a ``view confidence'' term that downweights detections with large angular offsets from the central field of view. The VL score is computed as following equations:
\begin{equation}\label{eq:VL}
\resizebox{0.9\linewidth}{!}{$
S_{\mathrm{VL}}(\mathbf{g})
\;=\;
\sum_{k=1}^{K}
\alpha_k\;
\exp\!\Bigl(
   -\tfrac{1}{2}
   \Bigl(\tfrac{\Delta\theta - \mu_k}{\sigma_k}\Bigr)^2
\Bigr)
\;\cdot\;
C\bigl(\Delta\theta\bigr),
$}
\end{equation}

\begin{equation}\label{eq:C_theta}
\resizebox{0.48\linewidth}{!}{$
C(\Delta\theta) = \cos^2\!\biggl(\frac{\Delta\theta}{(\theta_{\mathrm{fov}}/2)} \cdot \frac{\pi}{2}\biggr).
$}
\end{equation}
Finally, to ensure the metric remains bounded, the accumulated score \(S_{\mathrm{VL}}(\mathbf{g})\) is clipped to the range \([0, 1]\).

\paragraph{Curiosity Cues:}
We add two curiosity terms to guide navigation toward larger unexplored areas and prevent the system from repeatedly selecting faraway goals that offer only a marginal increase in VL score—thus reducing unnecessary back-and-forth movement which is extremely important for the large-scale environments like the hallway and the outdoor environments in our experiments.

\emph{(1) Distance Weighting.} We define:
\begin{equation}\label{eq:dist}
\resizebox{0.4\linewidth}{!}{$
S_{\mathrm{dist}}(\mathbf{g})
\;=\;
\frac{1}{1 + d(\mathbf{x}_r,\mathbf{g})},
$}
\end{equation}
so that nearer goals receive slightly higher scores. 
This factor is especially important on a real robot, as shorter travel distances can significantly reduce energy consumption while preventing needless wandering.
Although the distance term alone does not guarantee an optimal path, it helps prioritize goals that can be reached sooner.

\emph{(2) Unknown-Area Weighting.} We encourage curiosity-driven exploration by measuring how many unknown cells lie around \(\mathbf{g}\). First, we define the ratio of unknown as:
\begin{equation}\label{eq:ratio}
\resizebox{0.50\linewidth}{!}{$
\mathrm{ratio}(\mathbf{g})
\;=\;
\frac{\#(\text{unknown cells})}{\#(\text{total visited cells})},
$}
\end{equation}
where a local BFS radiates outward from \(\mathbf{g}\), counting how many cells are unknown versus how many are reachable (i.e., not blocked by obstacles). Larger \(\mathrm{ratio}(\mathbf{g})\) implies that moving to \(\mathbf{g}\) may reveal significant unknown space and get more information. 
To normalize this raw ratio, we define:
% To translate this raw ratio into a normalized weighting score, we apply an exponential mapping:
\begin{equation}\label{eq:unknown}
% \resizebox{0.70\linewidth}{!}{$
S_{\mathrm{unknown}}(\mathbf{g})=
1 -
\exp^{-k\cdot\mathrm{ratio}(\mathbf{g})},
% $}
\end{equation}
where $k$ is  an adjustable parameter that controls how rapidly the score increases from 0 toward 1.

\paragraph{Combined NeSy Score:}
The final scores of frontier-based goals combine these three components:
\begin{equation}\label{eq:NeSy}
\resizebox{0.91\linewidth}{!}{$
S_{\mathrm{NeSy}}(\mathbf{g})
\;=\;
\Bigl(
   w_{\mathrm{dist}}\;
   S_{\mathrm{dist}}(\mathbf{g})
   \;
   +\;
   w_{\mathrm{VL}}\;
   S_{\mathrm{VL}}(\mathbf{g})
   \;\cdot\;
   S_{\mathrm{unknown}}(\mathbf{g})
\Bigr),
$}
\end{equation}
where \(w_{\mathrm{VL}}\) and \(w_{\mathrm{dist}}\) are scalar weights. Note that the VL score is assigned when a point is initially detected, whereas the curiosity terms (distance and unknown-area weighting) are evaluated at goal selection time, when the current distance and unknown ratio can be calculated. For instance-based goals, these curiosity cues are omitted because their primary purpose is verification rather than exploration.

\subsection{Goal Selection \& Path Planning}

After computing scores, the system selects the optimal goal point by prioritizing high-confidence semantic matches over general exploration. Specifically, we iterate through all available \textit{instance-based target points} that exceed a minimum distance threshold \(\delta_{\mathrm{reached}}\). If any valid instance targets exist, the one with the highest \(S_{\mathrm{VL}}\) is selected immediately to drive verification. 
If no valid instance targets are available (or all are too close), the system falls back to \textit{frontier-based target points}. The frontier with the highest combined \(S_{\mathrm{NeSy}}\) score is chosen to maximize information gain. If no candidate exceeds \(\delta_{\mathrm{reached}}\), the robot holds its position. Once a goal is selected, we employ the FAR Planner \cite{yang2022far} to generate a collision-free path, guiding the robot to the target while dynamically avoiding real-time obstacles.

% \subsection{Path Planning with FAR Planner}
% Once an NeSy goal is selected, we hand off point-goal path planning to the FAR Planner \cite{yang2022far}, which represents obstacles polygonally and updates a visibility graph in real time. This enables efficient re-planning in partially unknown environments, often outperforming search- or sampling-based methods. A local planner refines FAR Planner’s waypoints into short-horizon velocity commands, ensuring swift reactions to new obstacles and seamless integration with VL-Nav’s zero-shot vision-language objectives.

%% file: sec/4_evaluation.tex
\section{Experiments}
\label{sec:experiments}

\subsection{Simulation Experiments: DARPA TIAMAT Challenge}
\label{sec:darpa_exp}

We first validate VL-Nav in the high-fidelity simulation environment of the DARPA TIAMAT Challenge (Phase 1). This benchmark is designed to test robotic reasoning under highly abstract and complex human instructions.

\paragraph{Baselines:}
We compare VL-Nav against four well-known baselines and two ablation variants:
closest frontier exploration, VLFM \cite{yokoyama2024vlfm}, SG-Nav \cite{yin2025sg}, ApexNav \cite{zhang2025apexnav}, VL-Nav w/o IBTP (ablation), and VL-Nav w/o Curiosity (ablation).

\paragraph{Environments \& Simulators:}
The evaluation covers four distinct scenarios simulated across two platforms:
\begin{itemize}
    \item \textbf{HabitatSim (Indoor):} The \textit{Apartment 1} and \textit{Apartment 2} environments are rendered in HabitatSim \cite{savva2019habitat} to evaluate navigation in daily residential settings.
    \item \textbf{IsaacSim (Outdoor):} The \textit{Camping Site} and \textit{Factory} environments are simulated in IsaacSim \cite{isaacsim} to leverage its advanced physics engine for unstructured settings.
\end{itemize}

\paragraph{Task Settings:}
Except for interaction, each environment consists of \textbf{8 abstract navigation tasks} requiring multi-step reasoning. To ensure reliability, we conduct 3 trials for all tasks in each environment. Crucially, some tasks require the robot to locate multiple target objects distributed across widely separated positions within a large-scale map. Instructions involve implicit semantics (e.g., ``it is raining'' $\rightarrow$ find umbrella), hierarchical searching (e.g., ``check the garage for tools''), and abstract social concepts (e.g., ``prepare for a black-tie party'' $\rightarrow$ find suit, shoes, and tie).

\noindent\textbf{Robot and Sensor Setup:}
The DARPA TIAMAT tasks require a Boston Dynamics Spot robot, equipped with five RGB-D cameras positioned to provide a comprehensive surround view for visual reasoning. 
% To strictly evaluate the reasoning and exploration logic, ground truth pose is provided during testing to decouple localization errors from navigation performance.

\begin{table*}[ht]
% \captionsetup{
%     width=\textwidth,
%     font=Smallfont,
%     labelfont=Smallfont,
%     textfont=Smallfont
%     }
\caption{Simulation experiment results on the navigation tasks of DARPA TIAMAT challenge phase 1 in 4 different scenarios with metrics Success Rate (SR) and Max Time Usage Ratio (MTUR).}
\label{SimResults}
\centering
\begin{tabular}{lcccc|cccc}
\toprule
\multirow{2}{*}{\textbf{Method}} & \multicolumn{4}{c}{\textbf{SR \% ($\uparrow$)}} & \multicolumn{4}{c}{\textbf{MTUR ($\downarrow$)}} \\
\cmidrule(lr){2-5} \cmidrule(lr){6-9}
 & Apartment 1 & Apartment 2 & Camping Site & Factory & Apartment 1 & Apartment 2 & Camping Site & Factory \\
\midrule
\textbf{Frontier Exploration}  
  & 8.3    & 8.3    & 0.0 & 0.0  
  & 0.958 & 0.884 & 1.000 & 1.000 \\

\textbf{VLFM} \cite{yokoyama2024vlfm}                  
  & 8.3    & 8.3    & 4.2    & 8.3  
  & 0.931 & 0.864 & 0.953 & 0.859 \\

\textbf{SG-Nav} \cite{yin2025sg}      
  & 0.0 & 4.2    & 0.0 & 8.3  
  & 1.000 & 0.973 & 1.000 & 0.901 \\

\textbf{ApexNav} \cite{zhang2025apexnav}      
  & 25.0 & 25.0 & 20.8 & 12.5
  & 0.817 & 0.795 & 0.828 & 0.861 \\
\midrule
\textbf{VL-Nav w/o IBTP}
  & 70.8 & 62.5 & \underline{62.5} & 58.3
  & 0.680 & 0.724 & \underline{0.731} & 0.762 \\

\textbf{VL-Nav w/o Curiosity}
  & \underline{79.1} & \underline{75.0} & 58.3 & \underline{66.7}
  & \underline{0.612} & \underline{0.635} & 0.793 & \underline{0.735} \\

\textbf{VL-Nav}               
  & \textbf{87.5} & \textbf{79.2} & \textbf{75.0} & \textbf{75.0}  
  & \textbf{0.562} & \textbf{0.591} & \textbf{0.647} & \textbf{0.679} \\

\bottomrule
\end{tabular}
\vspace{-6pt}
\end{table*}

\begin{figure}[t]
    \centering
    \includegraphics[width=1\linewidth]{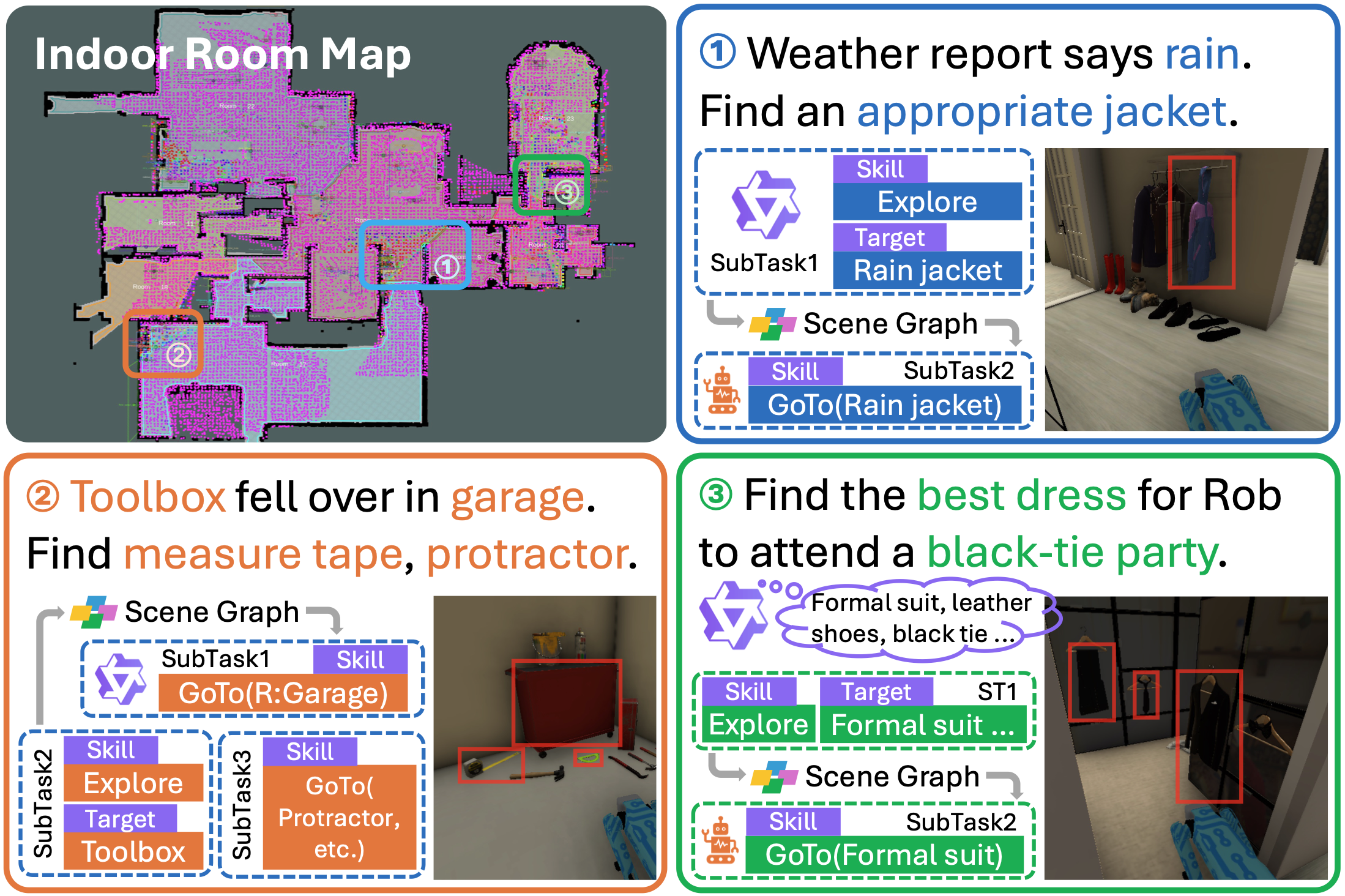} 
    % \vspace{-20pt}
    \caption{\textbf{Qualitative results in the DARPA TIAMAT Indoor-2 environment.} 
We demonstrate the system's capability to resolve three distinct types of complex instructions: 
(1) implicit semantic inference (weather $\rightarrow$ rain jacket); 
(2) hierarchical task decomposition (garage $\rightarrow$ tools); and 
(3) abstract concept grounding (black-tie party $\rightarrow$ formal suit).}
\label{fig:darpa_indoor}
    \vspace{-10pt}
\end{figure}

\begin{figure}[t]
    \centering
    \includegraphics[width=1\linewidth]{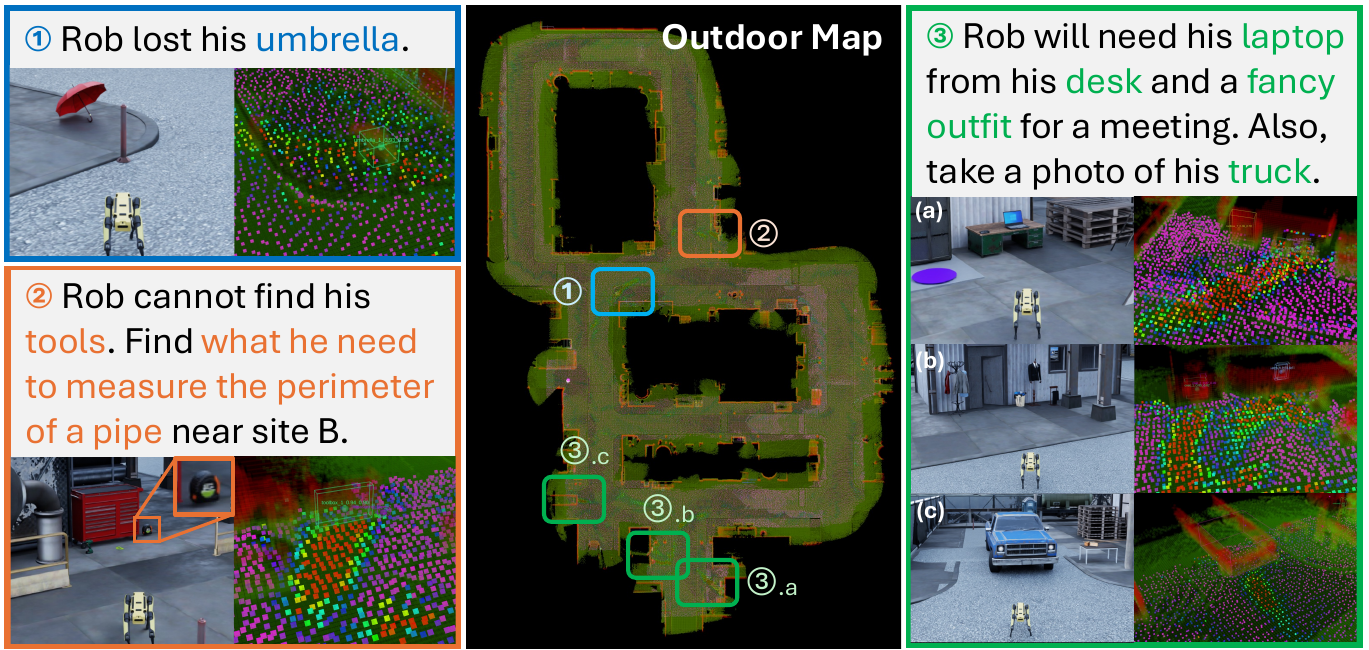} 
    % \vspace{-20pt}
    \caption{\textbf{Qualitative results in the DARPA TIAMAT outdoor factory environment.} 
We demonstrate robust navigation under unstructured conditions across three scenarios: 
(1) single-object retrieval; 
(2) implicit inference (measure the perimeter of a pipe $\rightarrow$ measuring tape); and 
(3) a challenging multi-objective task where the robot must locate three distinct targets scattered across different regions of the large-scale map. 
The colored point clouds in the insets visualize the value map guiding the exploration.}
\label{fig:darpa_outdoor}
\vspace{-18pt}
    
\end{figure}

\begin{figure*}[t]
    \centering
    \includegraphics[width=1\linewidth]{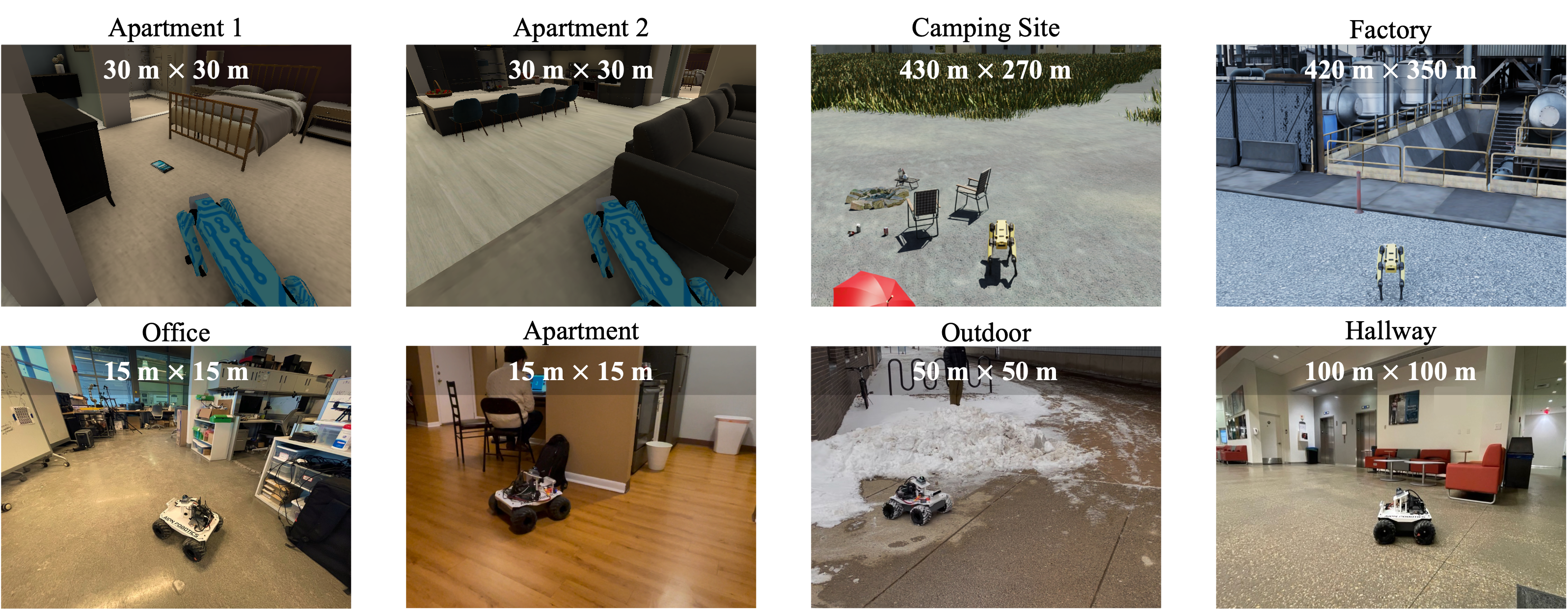} 
    % \vspace{-20pt}
    % \captionsetup{
    % width=\textwidth,
    % font=Smallfont,
    % labelfont=Smallfont,
    % textfont=Smallfont
    % }
    \captionsetup{
    width=\textwidth,
    font=Smallfont,
    labelfont=Smallfont,
    textfont=Smallfont
    }
    \caption{\textbf{Top:} Four simulation environments in the DARTPA TIAMAT Challenge. \textbf{Bottom:} Four  real-world experiment environments.}
    \label{fig:environments}
    \vspace{-15pt}
\end{figure*}

\begin{table*}[ht]
% \captionsetup{
%     width=\textwidth,
%     font=Smallfont,
%     labelfont=Smallfont,
%     textfont=Smallfont
%     }
\caption{Vision-language navigation performance in 4 unseen environments (SR and SPL).}
\label{SOTAResults_Real}
\centering
\begin{tabular}{lcccc|cccc}
\toprule
\multirow{2}{*}{\textbf{Method}} & \multicolumn{4}{c}{\textbf{SR \% ($\uparrow$)}} & \multicolumn{4}{c}{\textbf{SPL ($\uparrow$)}} \\
\cmidrule(lr){2-5} \cmidrule(lr){6-9}
 & Hallway & Office & Apartment & Outdoor & Hallway & Office & Apartment & Outdoor \\
\midrule
\textbf{Frontier Exploration}  
  & 40.0 & 41.7 & 55.6 & 33.3  
  & 0.239 & 0.317 & 0.363 & 0.189 \\

\textbf{VLFM} \cite{yokoyama2024vlfm}                  
  & 53.3 & 75.0 & 66.7 & 44.4  
  & 0.366 & 0.556 & 0.412 & 0.308 \\

\textbf{VL-Nav w/o IBTP}      
  & 66.7 & 83.3 & \underline{70.2} & \underline{55.6}  
  & 0.593 & 0.738 & 0.615 & \underline{0.573} \\

\textbf{VL-Nav w/o curiosity}      
  & \underline{73.3} & \underline{86.3} & 66.7 & \underline{55.6}  
  & \underline{0.612} & \underline{0.743} & \underline{0.631} & 0.498 \\

\textbf{VL-Nav}               
  & \textbf{86.7} & \textbf{91.7} & \textbf{88.9} & \textbf{77.8}  
  & \textbf{0.672} & \textbf{0.812} & \textbf{0.733} & \textbf{0.637} \\

\bottomrule
\end{tabular}
\vspace{-6pt}
\end{table*}

\subsection{Real-World Experiments}
\paragraph{Baselines:}
\label{sec:experimental_setting}
Since SG-Nav \cite{yin2025sg} and ApexNav \cite{zhang2025apexnav} cannot be deployed on the edge device,
we evaluate our approach in real-robot experiments against five methods: (1) closet frontier-based exploration, (2) VLFM \cite{yokoyama2024vlfm}, (3) VL-Nav without instance-based target points, (4) VL-Nav without curiosity terms, and (5) the full VL-Nav configuration. Because the original VLFM relies on BLIP-2 \cite{li2023blip}, which is too computationally heavy for real-time edge deployment, we use the same YOLO-World \cite{cheng2024yolo} and FastSAM \cite{zhao2023fast} models instead to generate per-observation similarity scores for VLFM. Each method is tested under the same conditions to ensure a fair comparison of performance.

\paragraph{Environments:}
We consider four distinct environments (shown in \fref{fig:environments}), each with a specific combination of semantic complexity (\textit{High}, \textit{Medium}, or \textit{Low}) and size (\textit{Big}, \textit{Mid}, or \textit{Small}). Concretely, we use a Hallway (\textit{Medium \& Big}), an Office (\textit{High \& Mid}), an Outdoor area (\textit{Low \& Big}), and an Apartment (\textit{High \& Small}).

\paragraph{Task Settings:}
We define nine distinct, uncommon human-described instances to serve as target objects or persons during navigation (e.g., ``tall white trash bin,'' ``there seems to be a man in white''). This results in 9 navigation tasks in total, and we conduct 3 to 5 trials for each task to robustly evaluate performance. This ensures that the robot must rely on NeSy reasoning to accurately locate these targets.

\paragraph{Real Robot Implementation:} 
Experiments are conducted on a four-wheeled Rover, with a Unitree Go2 quadruped used for the multi-floor demonstration. Both robots are equipped with a Livox Mid-360 LiDAR (tilted $23^{\circ}$) and an Intel RealSense D455 camera (tilted $7^{\circ}$). State estimation is handled by Super Odometry \cite{zhao2025resilient}, ensuring resilient robot state estimation throughout the mission. The Go2 is additionally equipped with a Unitree D1 arm and a RealSense D435 camera to enable object grasping. Computation is distributed: the real-time navigation stack optimized with pypose \cite{wang2023pypose} runs onboard on an NVIDIA Jetson Orin NX, while the NeSy Task Planner runs on a remote RTX 4090 device to host the Qwen3-VL-8B model \cite{bai2025qwen3}. This separation is effective as the task planner operates asynchronously, updating only when a new task is received or a subtask is completed.

\subsection{Main Results}
\label{sec:main_results}

We validate the proposed VL-Nav system through rigorous testing in both the high-fidelity DARPA TIAMAT simulation and diverse real-world deployments. Our evaluation employs distinct metrics suited for each domain: for simulation, we report Success Rate (SR) and Max Time Usage Ratio (MTUR), defined as the average execution time divided by the maximum time limit per task, to quantify time efficiency and completion reliability. For real-world experiments, we adopt SR and Success weighted by Path Length (SPL) following the VLFM convention \cite{yokoyama2024vlfm} to evaluate efficiency.

\paragraph{Simulation Results (DARPA TIAMAT)}
As reported in Table~\ref{SimResults}, VL-Nav achieves dominant performance with SRs ranging from 75.0\% to 87.5\%, whereas baselines struggle.
\begin{itemize}
    \item \textbf{Baseline Failures and High MTUR:} Standard approaches fail to resolve the abstract logic and multi-target sequencing required by the benchmark. Notably, \textbf{SG-Nav} \cite{yin2025sg} exhibits an MTUR near 1.0 across environments, frequently timing out. This is primarily because its reliance on LLM-based Chain-of-Thought reasoning at \textit{every navigation step} incurs prohibitive latency \cite{yin2025sg}. Similarly, \textbf{ApexNav} \cite{zhang2025apexnav} suffers from low success (SR $\approx$ 12.5\%--25.0\%) as it lacks the consistent symbolic memory to efficiently manage long-horizon and multiple goals located at different positions.
    \item \textbf{Qualitative Reasoning:} The system's ability to ground abstract concepts is visualized in Figures~\ref{fig:darpa_indoor} and~\ref{fig:darpa_outdoor}. In indoor settings (Fig.~\ref{fig:darpa_indoor}), VL-Nav successfully infers implicit semantics (e.g., ``weather'' $\rightarrow$ ``rain jacket'') and performs hierarchical decomposition. In unstructured outdoor settings (Fig.~\ref{fig:darpa_outdoor}), the projected value maps demonstrate how the system targets large landmarks to guide the search for smaller, distributed objects (laptop, outfit, truck) without getting lost in open space.
\end{itemize}

\paragraph{Real-World Results}
Physical deployments (Table~\ref{SOTAResults_Real}) confirm the system's robustness and efficiency.
\begin{itemize}
    \item \textbf{Efficiency (SPL):} VL-Nav achieves significantly higher SPL scores (e.g., 0.812 in Office) compared to Frontier Exploration (0.317) and VLFM (0.556). This metric confirms that VL-Nav does not merely find targets by chance but navigates via efficient, non-redundant paths, verifying that the neuro-symbolic planning translates effectively to real-world dynamics.
    \item \textbf{Sim-to-Real Transfer:} The consistent performance gap between VL-Nav and baselines across both simulation and real-world experiments highlights the system's strong generalization capabilities.
\end{itemize}

\paragraph{Ablation Analysis}
% We further analyze the contribution of specific components:
\begin{itemize}
    \item \textbf{Impact of IBTP (Verification):} The \textit{w/o IBTP} variant suffers most in semantically complex and cluttered environments, such as the \textit{Apartment} (Sim: 87.5\% $\rightarrow$ 70.8\%; Real: 88.9\% $\rightarrow$ 70.2\%). Without IBTP, the robot lacks the ``short-cut'' mechanism to verify tentative detections, leading to missed targets and lower SPL.
    \item \textbf{Impact of Curiosity (Exploration):} The \textit{w/o Curiosity} variant degrades primarily in large-scale, open environments, such as the \textit{Factory} and \textit{Outdoor} scenes (Sim Factory: 75.0\% $\rightarrow$ 66.7\%; Real Outdoor: 77.8\% $\rightarrow$ 55.6\%). This confirms that distance and unknown-area weighting are critical for preventing local minima and aimless wandering in expansive maps.
\end{itemize}

% \paragraph{Summary}
% The combined results demonstrate that while baselines like SG-Nav and ApexNav stumble on the computational and logical demands of abstract multi-target tasks, VL-Nav's neuro-symbolic architecture provides the necessary reasoning depth and execution efficiency to succeed in both high-fidelity simulation and real-world environments.

%% file: sec/6_conclusion.tex
\section{Conclusions and Future Work}
\label{sec:limitation_conclusion}

% \subsection{Limitations}
% \label{sec:limitation}
% Despite the promising results demonstrated by VL-Nav, several limitations remain. One major challenge is the system's ability to handle complex language descriptions. It struggles with spatial descriptions that contain hidden object references and objects described with specific textual annotations, which can affect navigation accuracy. 
% Another limitation lies in the reliance on manually defined thresholds for various navigation conditions like the lightning condition. These thresholds may not generalize well across different environments and scenarios, requiring further investigation into adaptive or learning-based approaches to threshold tuning.
In this paper, we presented VL-Nav, a neuro-symbolic framework that bridges the gap between abstract human instructions and robotic execution. By intertwining a NeSy Task Planner with a NeSy Exploration System, our approach effectively addresses the logical and spatial reasoning challenges inherent in large-scale vision-language navigation. Extensive validation across the DARPA TIAMAT challenge and diverse real-world experiments demonstrates the system's robustness, achieving overall success rates of 79.2\% and 86.3\%, respectively. These results confirm VL-Nav's capability to complete complex tasks in large-scale unseen environments.
Future work will focus on extending the symbolic memory to support temporal reasoning, enabling the robot to track moving targets and adapt to dynamic environmental changes evolving environments.